\documentclass{article}

\pdfoutput=1

\usepackage{arxiv}
\usepackage{booktabs}
\usepackage[utf8]{inputenc} 
\usepackage[T1]{fontenc}    
\usepackage{hyperref}       
\usepackage{url}            
\usepackage{tabularx}
\usepackage{booktabs}     
\usepackage{amsfonts}       
\usepackage{nicefrac}       
\usepackage{microtype}      
\usepackage{lipsum}		
\usepackage{graphicx}
\usepackage{natbib}
\usepackage{doi}
\usepackage{amsmath}
\usepackage{algorithm}
\usepackage{algpseudocode}
\usepackage{authblk}

\title{ Graph Agent: Explicit Reasoning Agent for Graphs }

\author[1,2]{Qinyong Wang}
\author[1]{Zhenxiang Gao}
\author[1]{Rong Xu}

\affil[1]{Center for Artificial Intelligence in Drug Discovery, Case Western Reserve University}
\affil[2]{Department of Computer and Data Sciences, Case Western Reserve University} 

\affil[ ]{\text {\{qxw225, zxg306, rxx\}@case.edu}}

\date{}

\hypersetup{
pdftitle={EXPLICIT REASONING ON KNOWLEDGE GRAPH},
}

\begin{document}
\maketitle
\begin{abstract}
 Graph embedding methods such as  Graph Neural Networks (GNNs) and Graph Transformers have contributed to the development of graph reasoning algorithms for various tasks on knowledge graphs. However, the lack of interpretability and explainability of graph embedding methods has limited their applicability in scenarios requiring explicit reasoning. In this paper, we introduce the Graph Agent (GA), an intelligent agent methodology of leveraging large language models (LLMs), inductive-deductive reasoning modules, and long-term memory for knowledge graph reasoning tasks. GA integrates aspects of symbolic reasoning and existing graph embedding methods to provide an innovative approach for complex graph reasoning tasks. By converting graph structures into textual data, GA enables LLMs to process, reason, and provide predictions alongside human-interpretable explanations. The effectiveness of the GA was evaluated on node classification and link prediction tasks. Results showed that GA reached state-of-the-art performance, demonstrating accuracy of 90.65\%, 95.48\%, and 89.32\% on Cora, PubMed, and PrimeKG datasets, respectively. Compared to existing GNN and transformer models, GA offered advantages of explicit reasoning ability, free-of-training, easy adaption to various graph reasoning tasks.
\end{abstract}


\begin{figure}
  \centering
  \includegraphics[scale=0.45]{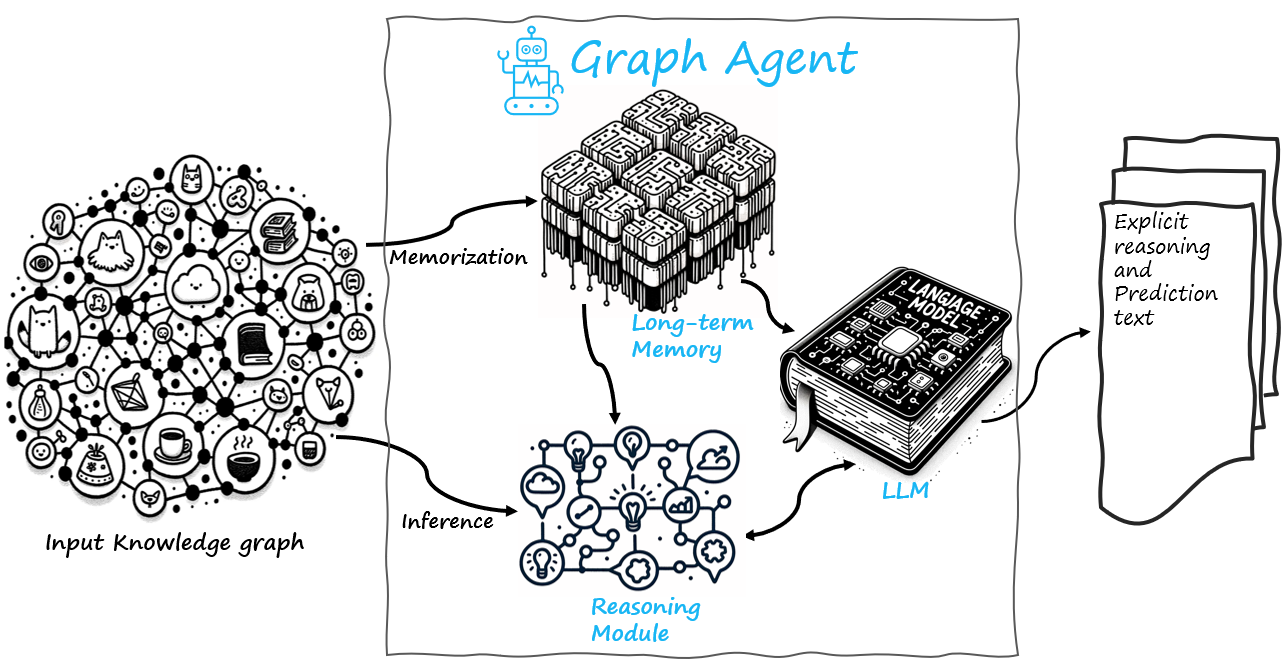}
  \caption{Overview of Graph Agent methodology}
  \label{fig:overview}
\end{figure}

\section{Introduction}

Knowledge graph \cite{hogan2021knowledge} has emerged as a pivotal structure for organizing and representing vast amounts of human knowledge in explicit form \cite{pan2023large}. Graph embedding methods, or predominantly Graph Neural Networks (GNNs) \cite{zhou2020graph} in recent years, have been used in capturing intricate graph structures and node features based on their local neighborhoods, making them adept at various graph reasoning tasks.

GNNs have demonstrated proficiency across various tasks, yet their  explainability of predictions remains a significant challenge \cite{dai2022comprehensive}. Such techniques, we termed "implicit reasoning methods," depend on entangled representations \cite{wang2019controllable}. GNNs utilize message passing to integrate information from adjacent nodes. However, during this process, both node features and the GNN kernels are numeric vectors, which are hard to interpret and understand by humans. This disadvantage poses a significant concern, especially in scientific discovery using graph data\cite{zeng2022toward, zhao2021knowledge}, where mere prediction outcomes are insufficient; a comprehensive understanding of the underlying rationale is imperative. Moreover, given that information in KGs possesses an explicit form \cite{pan2023large}, a rationale exists for pursuing explicit graph reasoning.

On the other hand, there are symbolic reasoning frameworks available. These might involve rule-based systems \cite{jamian2019rule, yao2019clinical} or using first-order logic \cite{belle2020symbolic, jane2019review} to reason with intricate data. Such reasoning processes are more transparent and human-interpretable. Hence, we term them "explicit reasoning methods". Yet, they come with their own set of problems. Rules within these systems are static, making adaptability to varied tasks challenging \cite{belle2020symbolic}. The current rule-based and symbolic reasoning methods could only work on a limited number of data sets. The intricate nature of heterogeneous graph data often proves too complex to be encapsulated solely by a simple logical system. These explicit logical reasoning methods often under-perform when benchmarked on various data sets \cite{belle2020symbolic}.

Large language models(LLM) demonstrate commendable reasoning capabilities \cite{bubeck2023sparks}, with their reasoning processes in natural language or symbolic language \cite{gao2023pal}. Recent investigations\cite{guo2023gpt4graph} have illuminated that LLMs can understand graph structures and analyze the encompassed information. Such ability paved the way for formulating graph reasoning algorithms with high efficacy and applicability. Researchers have employed LLMs as controllers or agents across a spectrum of tasks \cite{park2023generative}, such as software development \cite{hong2023metagpt} and robotic planning\cite{fan2022minedojo, shah2023lm}. However, the potential of LLM agent systems on complex graph reasoning is yet to be discovered.

We introduce the Graph Agent methodology that leverages the reasoning capabilities of LLMs and long-term memory\cite{zhong2023memorybank} for KG tasks. Given a graph data set, nodes or edges are embedded and stored in long-term memory. For each input link prediction or node classification sample, similar nodes or edges from the training set are fetched from this long-term memory. The LLM then undergoes a two-phase inductive-deductive reasoning process. During induction, the LLM is provided with similar nodes or edges, their neighbors, and associated labels from the prompt and concludes the rationale behind their labeling. In the deduction phase, the LLM incorporates these concluded reasons to reason and predict the presented sample. The underlying principle is similar to human cognitive processes: recalling similar past instances to inform decisions on unfamiliar problems. Such analogical reasoning, which widely exists in human cognition \cite{bartha2013analogy}, has previously been validated as effective in knowledge graph contexts \cite{yuan2023analogykb}. GA employs explicit reasoning, producing human-interpretable natural language outputs. Contrasting GA with GNNs, while GNNs retain learned patterns within graph convolution kernels, GA preserves it in textual format after inductive reasoning. Instead of the entangled representation utilized for message passing in GNNs, GA conveys neighbor information through prompts in natural language. Essentially, GA shares some mechanics with GNNs but in an explicit manner. The overview of GA is shown in Figure \ref{fig:overview}.

We evaluate GA's performance on node classification and link prediction datasets, notably Cora, PubMed, and PrimeKG data-sets, which are challenging to GNN-based approaches. Our findings indicate that GA excels in these tasks and offers enhanced prediction explainability compared to prior methods.

\section{Related Work}

\subsection{Graph neural networks and graph transformers}

A standard GNN \cite{zhou2020graph} is structured with layers that aggregate information from neighboring nodes. Message Passing Networks\cite{hamilton2017inductive} are a foundational framework for many GNNs. The core idea is to iteratively update node representations by "passing messages" between nodes. At each iteration, each node aggregates information (messages) from its neighbors. The node then updates its representation based on its previous state and the aggregated messages. The final node representations can be used for various tasks, such as node classification, graph classification, or link prediction. Graph Convolutional Networks (GCN) \cite{wu2019simplifying} are one of the most popular and effective GNN architectures. 
Graph Attention Network (GAT) \cite{velickovic2017graph}introduces attention mechanisms \cite{vaswani2017attention} to the world of GNNs. Standard GNNs, designed for homogeneous graphs, may not be optimal for Heterogeneous Graph. Heterogeneous GNNs are designed to handle multiple node and edge types. They often involve multiple relation-specific aggregation functions. An example is the Relational Graph Convolutional Network (R-GCN) \cite{schlichtkrull2018modeling}, which uses different weight matrices for different relation types. There are also other GNNs such as Heterogeneous Graph Transformer \cite{hu2020heterogeneous}, Hypergraph Convolution and Hypergraph Attention \cite{bai2021hypergraph}. Transformer models were adapted to graph-structured data by covert both nodes and edges as tokens \cite{min2022transformer}. As previously discussed, these methods exhibit a deficiency in the explainability of predictions.

\subsection{Generative language model agents}

Recently, there has been a growing interest in enhancing LLMs with additional tools, memory, and sophisticated reasoning frameworks. Techniques such as the "Chain of Thoughts," \cite{wei2022chain} "Self-Consistency," \cite{wang2022self}and "Tree of Thoughts" \cite{yao2023tree} have been introduced to boost the reasoning capabilities of LLMs. A common practice among LLMs is the utilization of vector databases to maintain their long-term memory \cite{wang2023augmenting}, which is crucial for retaining comprehensive graph entities. Furthermore, LLMs are now being trained to use tools, with notable developments like Toolformer \cite{schick2023toolformer}, Visual Programming \cite{gupta2023visual}, and GeneGPT \cite{jin2023genegpt} leading the way. These advancements have significantly amplified the competencies of LLMs, leading to a surge in research exploring LLMs as intelligent agents\cite{liu2023agentbench}. In experimental setups, these agents are immersed in virtual environments, allowing researchers to observe and analyze their behaviors \cite{park2023generative}. Remarkably, LLMs have even been employed to simulate a software development \cite{hong2023metagpt}. It would be interesting to see how LLM agents behave in complex graph reasoning tasks.

\subsection{Graph reasoning with LLMs}

In recent advancements in graph-based reasoning, the synergistic combination of LLMs with graphs has showcased enhanced performance compared to the exclusive GNNs. Two primary methodologies have been identified in the realm of LLM-integrated graph techniques \cite{chen2023exploring}:

\textbf{LLM as an Augmentor}: LLMs are pivotal in augmenting graph data. They are adept at generating text contextually related to the graph nodes. This results in the enrichment of each node with additional textual features, thereby amplifying the depth and quality of information associated with each node. TAPE leveraged the capabilities of ChatGPT to enhance text-attributed graphs, demonstrating state-of-the-art performance in node classification tasks
 \cite{he2023explanations}. LLMs exhibit potential as annotators, enhancing performance in graph tasks where labels are absent \cite{chen2023label}. 

\textbf{LLM as a Predictor}: Deploying LLMs as predictors has also been a significant stride forward. By feeding the LLM with information about a node and its neighboring nodes, it is feasible to anticipate the class of the given node or infer the likelihood of a link existing between two nodes \cite{ye2023natural}. GraphText-ICL used a graph-to-text encoder and leveraged the In-context Learning(ICL) ability of LLM for node classification \cite{zhao2023graphtext}. Additionally, integrating LLMs with graph structures for fine-tuning has shown to have refined outcomes in tasks such as Substructure Counting and shortest path identification \cite{chai2023graphllm}.

Previous studies have investigated fine-tuning LLMs for graph reasoning tasks \cite{ye2023natural,zhao2023graphtext}, and have explored the utilization of advanced reasoning techniques to enhance the graph reasoning capabilities of LLMs. However, fine-tuning LLMs for specific tasks can be computationally intensive and time-consuming, posing challenges when adapting them to diverse datasets. Furthermore, approaches centered on prompt engineering have yielded suboptimal results compared to training-based methods \cite{zhao2023graphtext}.

\begin{figure}
  \centering
  \includegraphics[scale=0.5]{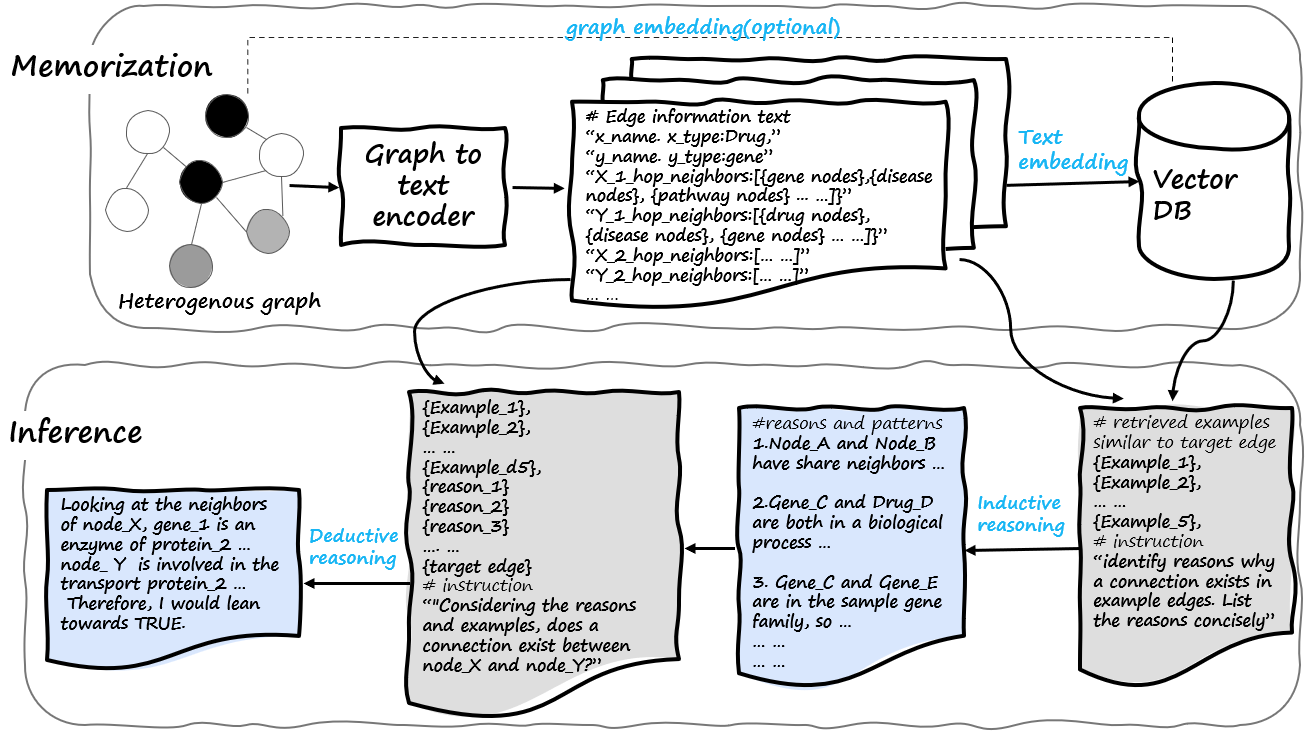}
  \caption{Workflow of the proposed Graph Agent methodology in link prediction task with an information-centered view. Graph Agent first memorized the train set of an input graph, then inferenced on the test set without model training. Cosine similarity was used to similar edges for inductive reasoning, then predictions and explanation was generated after deductive reasoning. Text boxes with gray backgrounds indicate input text for LLM; blue background indicate output text from LLM.}
  \label{fig:methodology1}
\end{figure}

\subsection{Graph to text encoder and node sampling}
To feed structured graph data to LLMs, we would need covert sub-graphs to text sequences. The efficacy of LLMs in graph reasoning tasks was influenced by the encoding method \cite{zhao2023graphtext}. Our study leveraged an encoding strategy optimized for computational efficiency.

For a given node \(v\) with attributes \(A\) and n-hop neighbors \(N_h\), the encoding process integrated the node's attributes \(\{a_n \in A\}\) and sampled information from its n-hop neighbors \(\{n \in N\}\) using a sampling function denoted as \(f_{\text{sample}}()\). The formulation of the node's encoder function is as follows:

\[
\text{encoder}(v, A, N) = [\text{{"node:"}}, v, \text{{"attributes:"}} A, \text{{"n-hop-neighbours: "}} [f_{\text{sample}}(N_h)]]
\]

For graph edges, represented by vertices \(x\) and \(y\), the encoding encompassed the attributes of the two edge nodes (\(A_x, A_y\)) and information pertaining to the n-hop neighbors for each vertex (\(N_x, N_y\)). The edge encoder function is as follows:

\begin{align*}
\text{encoder}((x, y), (A_x, A_y), N_x, N_y) = [ & \text{{``edge: ''}}, (x, y); \\
& \text{{``attributes: ''}}, (A_x, A_y); \\
& y\text{{-``n-hop-neighbours: ''}} [f_{\text{sample}}(N_{y_h})], \\
& x\text{{-``n-hop-neighbours: ''}} [f_{\text{sample}}(N_{x_h})]]
\end{align*}

This encoder intentionally omitted the interconnections among the neighboring nodes, which might affect the performance due to information loss. Including these connections would substantially increase the text length. It would prolong inference time and risk the LLM being overwhelmed by over-complicated graph structures.

Our experiments underscored the pivotal role of information sampling in optimizing GA, especially within graphs with dense connectivities. For instance, nodes in biomedical knowledge graphs or individuals in social networks often have connections exceeding hundreds of edges, posing a risk of overloading information beyond the working memory \cite{bubeck2023sparks} of LLMs, or exceeding the maximum context length of LLMs.

To mitigate this, we leveraged a sampling technique based on node degrees. For heterogeneous graphs, we computed the average degree \(D_{\text{avg}}\) for each category of nodes. Subsequently, a node's importance was quantified as the ratio of its degree to \(D_{\text{avg}}\) pertinent to its category. This relative importance metric guided the selective encoding of neighbor information, specifically incorporating only the top \(k\) most significant nodes, where \(k\) was tailored to the task, dataset, and LLM in use. The sampling function is formalized as:

\[
f_{\text{sample}}(N) = \text{select\_top\_k}\left(\left\{\frac{\text{degree}(n)}{D_{\text{avg, type}(n)}} : n \in N\right\}, k\right)
\]

where \(\text{degree}(n)\) denotes the degree of node \(n\), The function \(\text{select\_top\_k}\) is an operation selecting the \(k\) nodes with the highest node importance.

While more advanced sampling and encoding techniques could be used, our research prioritized the initial development of Graph Agent. Consequently, we adopted a straightforward method to facilitate implementation.
\subsection{Long-term memory}
Given a graph task, the first step of GA was memorization of the graph. Compared with other methods with a training phase, the train set was not used for back-propagation training; instead, Graph Agent embedded all training samples and stored them in a vector database. During inference on test samples, we retrieved similar samples from the long-term memory.

We leveraged two methods to embed samples. The first was language model embedding, and the second was GNN embedding. In language model embedding, for each sample, we used the graph to text encoder previously discussed and passed the output text of the encoder to an embedding-optimized language model. In GNN embedding, we trained a GNN on the training data set and stored the node embedding in a vector database. For edge embedding, we simply contacted the embedding of two nodes. We used the Cosine similarity of embedding to retrieve similar node or edge examples. 

\subsection{Inductive reasoning}
During the inference phase, GA initiated a retrieval of analogous examples from its long-term memory for a target sample; each example was denoted as \texttt{example\_n}. The aggregation of these examples formed a structured prompt augmented with task instructions. The prompt was formulated as follows: "Given the provided examples and your existing knowledge, identify reasons why example nodes are categorized as labeled or why a connection exists in example edges. List the reasons concisely."

Selecting these analogous examples was critical, as the LMM sought patterns or commonalities within the examples. This mechanism, termed \emph{explicit learning}, where the learned patterns were in the output text.
\subsection{Deductive reasoning}
In the deductive reasoning phase, all selected examples, outcomes from inductive reasoning, and the designated sample were integrated into a comprehensive prompt, and the prompt was followed with instructions and a question. For node classification, the LLM was queried with a crafted question: \textit{"Given the reasons and examples, determine the type of \texttt{node\_a} from the following options: [options...], think step by step then choose one of the options"} 

Similarly, for link prediction, the LLM was instructed: \textit{"Considering the reasons and examples, does a connection exist between \texttt{node\_a} and \texttt{node\_b}? think step by step, and choose either TRUE or FALSE."}

The LLM would respond with explicit reasoning in natural language and a definitive prediction. This output represented transparent, human-interpretable logic, demonstrating how GA leveraged patterns discerned through inductive reasoning to inform their predictions. The Deductive reasoning phase could be considered a form of the chain of thought method \cite{wei2022chain}. The difference was the examples were dynamically generated and tailored to the designated sample.

\section{Experiments}

\subsection{Node Classification}
Our experimentation with GA encompassed node classification tasks juxtaposing its efficacy against previous methodologies. A direct comparison between GA and GNNs is inequitable, primarily because text-attributed graph\cite{he2023explanations} node classification could be transformed into document classification \cite{minaee2021deep}, an LLM innately excels. Consequently, this section aims at establishing a robust benchmark against preceding methods.

\textbf{Datasets}
Conscious of the high inference costs and long inference time of LLMs, our study utilized comparatively smaller datasets—specifically, the widely recognized Cora and PubMed graph datasets \cite{yang2016revisiting}. These selections facilitated comparisons with state-of-the-art techniques but also adhered to manageable sizes, with Cora comprising 2,706 nodes and PubMed encompassing 19,711. Consistent with previous works, we implemented a 60\%/20\%/20\% partitioning for train/validation/test sets. Our experiments engaged versions of Cora and PubMed retaining their original textual data.

\textbf{Baselines}
Our performance assessment of GA involved juxtapositions with preceding GNN models, transformer models, and those LLM-related methods. The GNN models include GCN\cite{wu2019simplifying}, GAT\cite{velivckovic2017graph}, RevGAT\cite{li2021training}, etc. We also compared with Transformer-based graph learns, including CoarFormer \cite{kuang2021coarformer}, Graphormer\cite{ying2021transformers}, and GT \cite{dwivedi2020generalization}. 

More importantly, we compared with graph algorithms that leveraged LLMs in recently published papers\cite{ye2023natural}, including IntructGLM, TAPE\cite{he2023explanations}, and GraphText-ICL\cite{zhao2023graphtext}.


\textbf{Implementation Details}
For our experiments, we employed the gpt-4-0613 model as the LLM backend, and used embedding-ada-002 model for graph text embedding. Our methodology included sampling the top 8 neighboring nodes. The prompts for the target nodes comprised the title, abstract, authors, and keywords, while for neighboring nodes, we confined the information to the title, authors, and node type label. We masked the labels of the target node within the prompts since similar node examples were often the neighbors of the target node and revealed labels of the target node. After memorizing the training dataset, GA directly inferred the test dataset without training.

\textbf{Results}
GA outperformed GNN and transformer models, achieving state-of-the-art results on the Cora and PubMed datasets. It demonstrated superior accuracy, attaining 95.48\% on PubMed— the highest in this category—and 90.65\% on Cora, ranking second only to InstructGLM. Notably, both TAPE and InstructGLM require a training phase, highlighting GA's efficiency as it yields competitive results without the necessity of model training. Furthermore, when compared against the free-of-training method GraphText-ICL, which also utilized GPT-4, GA observed an increase of approximately 20 points in accuracy.

\begin{table}
    \centering
    \begin{tabular}{lccc}
    \hline
   & & Cora (Acc) & PubMed (Acc) \\
    \hline
    GAT & \cite{velivckovic2017graph} & 76.70  & 83.28 \\
    GraphSAGE  &\cite{hamilton2017inductive} & 86.58  & 86.85 \\
    GCN & \cite{wu2019simplifying} & 87.78  & 88.90 \\
    RevGAT  &\cite{li2021training} & 89.11  & 88.50  \\
    ACM-Snowball-3  &\cite{luan2022revisiting} & 89.59  & 90.96  \\
    ACM-GCN+  &\cite{luan2022revisiting} & 89.75  & 91.44  \\
    \hline
    Graphormer  &\cite{ying2021transformers} & 80.41  & 88.75  \\
    GT&\cite{dwivedi2020generalization}& 86.42  & 88.24 \\
    CoarFormer  &\cite{kuang2021coarformer} & 88.69  & 89.75  \\
    \hline
    InstructGLM & \cite{ye2023natural} & \textbf{90.77}  & 94.62  \\
    TAPE & \cite{he2023explanations} & 89.30 & \textbf{95.30} \\
    GraphText-ICL & \cite{zhao2023graphtext} & 68.3 & \-- \\
    \hline
    Graph Agent & & \textbf{90.65} & \textbf{95.48} \\
    \hline
    \end{tabular}
    \caption{Results on Cora and PubMed node classification}
    \label{tab:results}
\end{table}

\subsection{Link Prediction}

In pursuit of real-world applicability, we explored drug-gene link prediction within a heterogeneous biomedical knowledge graph. The interrelations of genes, drugs, diseases, and biological processes were complex, which made it an ideal task for us to test the comprehensive graph reasoning ability of GA. Secondly, as stated previously, the inference cost of GA was high. It would make more sense to test GA for high-value graph reasoning tasks. 

\textbf{Dataset}
We adopted the Precision Medicine Oriented Knowledge Graph(PrimeKG) dataset \cite{chandak2023building}, recognizing it as one of the latest and most complex biomedical graphs available. Our primary focus centered on the prediction of drug-gene edges. As delineated earlier, LLMs were bound by input length.  Consequently, we confined our attention to certain node and edge types; we used a subset of PrimeKG node types, including drugs, genes, biological processes, pathways, and diseases. Only edge types that interconnected these node types were taken into consideration. The filtered version of PrimeKG utilized for our analysis comprised 2,085 drug nodes, 19,001 gene nodes, 7,161 biological process nodes, 1,625 pathway nodes, and 2,658 disease nodes. This configuration encapsulated a complex network with a total of 954,438 edges of various types. Within PrimeKG, 20,417 drug-gene edges were identified, indicating existing associations. An equivalent number of non-associated drug-gene pairs were randomly generated and reintegrated into the dataset. These newly created links were labeled as negative, contrasting with the original positive associations. The data set was partitioned into 80\%, 10\%, and 10\% segments for training, validation, and testing.

\textbf{Baselines}
We benchmarked GA's performance against established GNNs and prompt-engineering methods. Comparisons were drawn with GCN\cite{wu2019simplifying} and Heterogeneous graph attention network(HGAT)\cite{wang2019heterogeneous}, both in isolation and in conjunction with text augmentation. We also compared GA with simple asking and chain-of-though\cite{wei2022chain} prompt methods. 

\textbf{Evaluation Metrics}
A notable limitation of LLMs in link prediction is their binary response format, wherein they can only output labels such as "True" or "False". Consequently,  metrics like mean reciprocal rank (MRR) and the hit rate for the top k candidates (hits@K) are incompatible with our specific scenario \cite{pan2023large}. Given this constraint, our evaluation adopted precision, recall, the F1 score, and the accuracy of the positive edges. 

\textbf{Implementation Details}
For comparative analysis, we trained 2-layer GCN and HGAT models on the training dataset. We also used the graph embedding from GNNs for retrieving similar edges in addition to language model embedding.  Capitalizing on PrimeKG's text-associated nodes, we also integrated text embedding—generated from text-embedding-ada-002 model —as initial node embedding in GNN training, which was a text-augmented training.  We used both Gpt-4 and LLaMa2-70B-chat-hf as our LLM backend. Our sampling strategy prioritized the top 15 neighbor nodes for GPT-4, and the top 5 neighbor nodes for LLaMa2-70B, since LLaMa had much fewer parameters and could only process a smaller local graph. The prompt included only node names and types, excluding other node attributes. Similar to node classification, edge examples could contain the targeted edge label. We only leveraged edge examples that had different nodes with targeted edges. In our approach, we extracted three analogous positive edge instances from long-term memory and arbitrarily selected two negative edges for inclusion in the prompt. We also explored alternative methodologies for comparison: the Simple Ask approach and the 5-shot Chain-of-Thought (COT) \cite{wei2022chain} technique. The Simple Ask method involved presenting the LLM with text after graph-to-text encoding and straightforwardly inquiring about the existence of a connection. The 5-shot COT, similar to the previous GraphText-ICL strategy, employed a consistent set of three positive and two negative edge examples.

\textbf{Results}
Table 2 illustrates that GA outperformed competing methods in link prediction, achieving an F1 score of 0.889 and an accuracy of 0.893. Compared to the 5-shot COT's accuracy of 0.803, GA enhanced GPT-4's graph reasoning capabilities in the nearly 10-point increase in both F1 and accuracy metrics. Notably, the 'Simple Ask' method yielded a mere 0.196 recall, suggesting GPT-4's initial inclination to deny the existence of most drug-gene associations. Minor improvements were witnessed with text augmentation for GNNs, which aligned with findings from prior studies\cite{chen2023exploring}. Furthermore, the efficacy of COT methods was confirmed, with a 20-point surge in accuracy over the 'Simple Ask' approach.

\begin{table}[ht]
    \centering
    \caption{Results on PrimeKG drug-gene link prediction}
    \label{tab:my_label}
    \begin{tabularx}{\linewidth}{lXcccc}
        \toprule
        & PrimeKG & \textbf{Precision} & \textbf{Recall} & \textbf{F1} & \textbf{Accuracy} \\
        \midrule
        \textbf{GNN methods} & HGAT & 0.831 & 0.826 & 0.829 & 0.837 \\
        & HGAT with text augmentation & 0.836 & 0.842 & 0.839 & 0.846 \\
        & GCN & 0.826 & 0.838 & 0.832 & 0.832 \\
        & GCN with text augmentation & 0.830 & 0.842	 & 0.836 & 0.837 \\
        \addlinespace
        \hline
        \textbf{LLM predictors} & Simple ask + Graph2text encoder& \textbf{0.964} & 0.196 & 0.325 & 0.599 \\
        & 5-shot COT + Graph2text encoder & 0.962 & 0.673 & 0.792 & 0.803 \\
        \addlinespace
        \hline
        \addlinespace
        & \textbf{Graph Agent} & 0.926 & \textbf{0.854} & \textbf{0.889} & \textbf{0.893} \\
        \bottomrule
    \end{tabularx}
\end{table}

\textbf{Ablation Results} Our ablation study, summarized in Table 3, evaluated the influence of n-hop information and the specific embeddings and LLM utilized. Contrary to expectations, incorporating more n-hop information into GA diminished performance: 1-hop, 2-hop, and 3-hop configurations yielded accuracies of 0.8902, 0.8202, and 0.774, respectively. We found that the LLM was significantly affected by the shared neighbor pattern, with more neighbors for each drug and gene node. LLMs found more shared neighbors, leading the LLM to think there were associations. However, this is not true for many biomedical knowledge graphs. For example, a common gene could be associated with many genes and biological processes, and within 2-hop of this common gene, many nodes do not have associations. Furthermore, leveraging GNN embeddings for similar edge example retrieval proved effective, with an accuracy of 0.869, surpassing the 0.846 achieved by HGAT. It showed we could use GA to enhance existing GNN methods. However, GNN embedding under-performed compared with language model embedding, which could be due to our GNN model was not refined for information retrieval. Testing with LLama2-70B had a precision of only 0.5464.  Hallucination of LLama2 was frequently observed, and it stated most drugs and genes had associations. Our explanation was that LLama2, with only 70B parameters, had not yet developed the emergent graph reasoning ability\cite{wei2022emergent} required for a complex biomedical graph.

\begin{table}
    \centering
    \begin{tabular}{lcccc}
    \hline
    & Precision & Recall & F1 & Accuracy \\
    \hline
    3-hop  with LM embedding & 0.698 & 0.955 & 0.807 & 0.774 \\
    2-hop  with LM embedding & 0.730 & 0.932 & 0.821 & 0.820 \\
    1-hop  with LM embedding (proposed) & 0.926 & 0.854 & 0.889 & 0.893 \\
    1-hop  with GNN embedding & 0.958 & 0.767 & 0.852 & 0.869 \\
    1-hop  with LM embedding and LLama2 70B & 0.546 & 0.956 & 0.696	 & 0.565 \\
    \hline
    \end{tabular}
    \caption{Ablation test results}
    \label{tab:ablation}
\end{table}

\textbf{Qualitative analyses}
In the link prediction experiments, Graph Agent, empowered by GPT-4, demonstrated commendable reasoning capabilities. A manual review of reasoning instances revealed GPT-4 had hallucinations with uncommon genes and diseases. We found GPT-4 was confused with genes within the same family, which often shared similar names and biological processes. Since those genes share similar functions, they usually have links with the same drugs, so even if the thinking processing was wrong, the prediction was correct. This process correctness was hard to evaluate \cite{bubeck2023sparks}, so we could not provide a quantitative score for the reasoning quality. Despite the factual error, GA was good at identifying common reasons for drug-gene associations, such as shared neighbors, key biological processes, and pathways. This behavior could give researchers insights during drug discovery.

\section{Discussion and limitations}
 Deep learning has frequently been criticized as a "black box". Now we have seen this very "black box" offer compelling reasoning and elucidations \cite{bubeck2023sparks}. This poses a contemplative question: \textbf{can we trust explanations derived from black box models?} Addressing this quandary on philosophical grounds remains elusive. Our analysis of many graph reasoning outcomes produced by the LLM, indeed affirms a commendable quality of reasoning. Future studies could investigate the reliability of LLM reasoning. 

 The computational intensity of LLMs translates to formidable latencies and costs, rendering the current GA impractical for large-scale graph reasoning. A potential solution could lie in a hybrid system, wherein the LLM is harnessed exclusively for hard samples or high-value cases. Explorations could also go towards determining if smaller fine-tuned LLM can emulate the efficacy of its larger version of LLM. This, however, surfaces additional complications, including curating the dataset and methodology for fine-tuning and subsequently ascertaining the generalizability of the fine-tuned graph LLMs.

Our current Graph Agent methodology encounters limitations regarding information coverage and flow. Unlike multi-layered GNNs, which facilitate good node coverage and employ neural networks to regulate inter-node information flow, GA is constrained by its reliance on sampled local graphs and naive sampling methods for information flow control. Future research directions could be the exploration of advanced information retrieval and control techniques to enhance GA's efficacy.

An additional limitation pertains to the issue of redundant reasoning. GA often uncovers rationales that are applicable across multiple samples, rendering the repetitive inductive reasoning for each instance. Subsequent research might explore the development of reasoning modules with higher efficiency.

Our approach did not address the hallucination problem of LLMs \cite{rawte2023survey}. While mitigating this phenomenon typically involves empowering agents with factual information retrieval tools, incorporating such tools within GA was avoided to prevent data leakage. However, for practical implementations of GA in open-world applications, the integration of tools is an indispensable consideration, underscoring an avenue for subsequent studies.

While considerable scholarship has been dedicated to generative agents in robotics, software development, and human-like conversation \cite{wu2023autogen}, a void exists in discussions centered on agents with graph reasoning capabilities. Knowledge Graphs, representing structures of human knowledge, remain important. Efficacious reasoning atop KGs has significance. Our current Graph Agent only exhibits capacities for shallow reasoning on knowledge graphs, thereby accessing the surface layers of graph information. There exists an exigency for concerted efforts to formulate graph reasoning datasets and train graph foundation model \cite{liu2023towards} for graph agents. The development of evaluative metrics that truly reflect the logical and factual correctness of graph reasoning remains a challenge.

\section{Conclusion}
Existing implicit graph reasoning methods lack explainability. To address this issue, we proposed the Graph Agent methodology with long-term memory and an inductive-deductive reasoning module. Graph Agent has demonstrated high efficacy and transparency of predictions within Knowledge graphs. Beyond the promising results, the uniqueness of our approach lies in addressing the longstanding explainability challenge in graph reasoning. As a first-of-its-kind framework, the current Graph Agent implementation is naive and primitive. Future studies could investigate advanced Graph Agents that can truly voyage and learn in human knowledge networks.

\bibliographystyle{unsrtnat}
\bibliography{references}







\end{document}